\documentclass[sigconf,nonacm]{acmart}
\usepackage{booktabs}
\usepackage{multirow}
\usepackage{tabularx}
\usepackage{array}
\usepackage{graphicx}
\usepackage{subcaption}
\usepackage{algorithm}
\usepackage{algorithmic}
\AtBeginDocument{%
  }

\setcopyright{none}
\begin{document}

\title{The Devil Is in the Leakage: A Disentangled Dual-Purification Framework for High-Fidelity Hairstyle Transfer}

\author{Jijie Li}
\email{lijijie2024@ia.ac.cn}
\affiliation{%
  \institution{SAI, UCAS; MAIS, CASIA}
  \city{Beijing}
  \country{China}
}

\author{Jiankuo Zhao}
\email{zhaojiankuo2024@ia.ac.cn}
\affiliation{%
  \institution{MAIS, CASIA; SAI, UCAS}
  \city{Beijing}
  \country{China}
}

\author{Xiangyu Zhu}
\authornote{Corresponding author.}
\email{xiangyu.zhu@ia.ac.cn}
\affiliation{%
  \institution{SAI, UCAS; MAIS, CASIA}
  \city{Beijing}
  \country{China}
}

\author{Zhen Lei}
\email{zhen.lei@ia.ac.cn}
\affiliation{%
  \institution{MAIS, CASIA; SAI, UCAS}
  \city{Beijing}
  \country{China}
}
\affiliation{%
  \institution{CAIR, HKSIS, CAS}
  \city{Hong Kong}
  \country{China}
}
\affiliation{%
  \institution{SCSE, FIE, M.U.S.T}
  \city{Macau}
  \country{China}
}

\begin{abstract}
Hairstyle transfer aims to synthesize a photorealistic portrait by transplanting the hairstyle from a reference image onto a source subject, while preserving the source's identity. While recent large-scale foundation models exhibit remarkable generative capabilities, they struggle with the zero-shot disentanglement required for such precise local editing, inherently entangling the reference hairstyle with its original facial identity and pose. To address these limitations through structural decomposition, a standard pipeline for hairstyle transfer typically decouples the process by first generating a ``bald'' image from the source and extracting identity-agnostic hairstyle features from a reference, fusing them to produce the final result. However, this methodology is frequently prone to several types of artifacts, including identity inconsistency and mismatched hair geometry. In this paper, we demonstrate that these artifacts stem from a more fundamental issue, which we term the leakage problem. This leakage is twofold: First, Identity Leakage in Hairstyle occurs when hairstyle features remain entangled with the reference's identity and pose. Second, Flaw Leakage in Bald arises when subtle geometric flaws left in the ``bald'' image are propagated into the synthesized hairstyle.  To address these issues, we propose the Dual-Purification Framework (DPF), which integrates two complementary purification strategies. The Adversarial Hairstyle Purification (AHP) module explicitly purges identity information from hairstyle features by adversarially minimizing hairstyle–bald mutual information. Concurrently, the Contrastive Geometric Purification (CGP) module introduces a contrastive objective that penalizes the model's reliance on these geometric flaws in the ``bald'' image, thereby suppressing the Flaw Leakage in Bald. By explicitly mitigating both components of leakage, DPF achieves state-of-the-art performance in high-fidelity, identity-preserving hairstyle transfer. 
\end{abstract}


\begin{teaserfigure}
    \includegraphics[width=\linewidth]{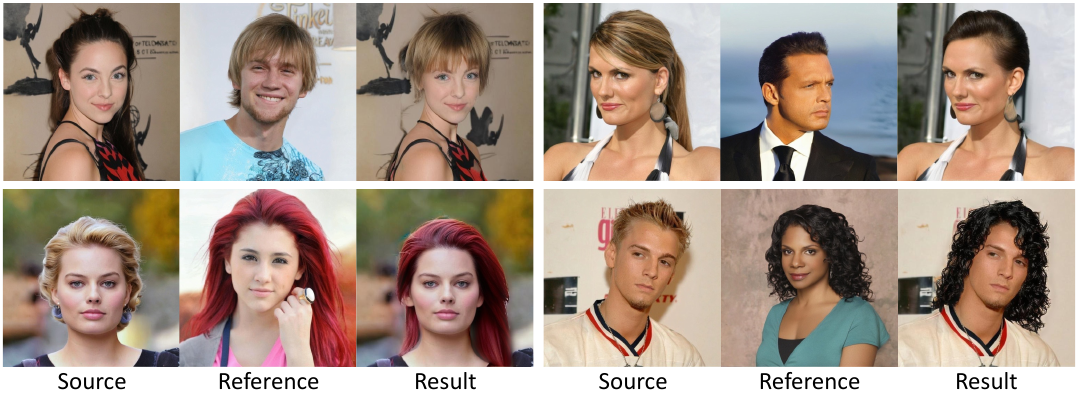}
    \caption{
    \textbf{High-fidelity hairstyle transfer by our proposed Dual-Purification Framework (DPF).} Our method successfully transfers complex hairstyles across subjects with significant variations in pose, gender, and appearance. DPF preserves the source identity with high fidelity and generates geometrically aligned, photorealistic results.
    }
    \label{fig:teaser}
\end{teaserfigure}


\maketitle
\begin{figure}[t]
  \centering
  \includegraphics[width=\columnwidth]{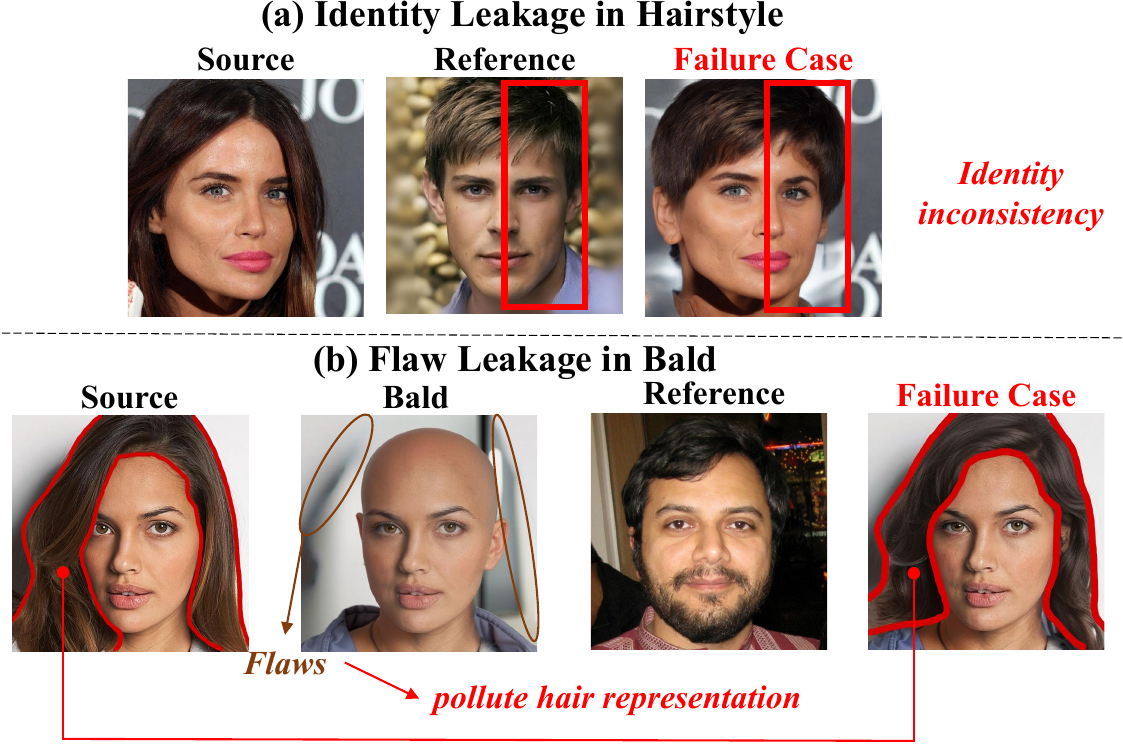}
 \vspace*{-5mm}
    \caption{
    \textbf{Illustration of the two factors of leakage in hairstyle transfer.}
    \textbf{(a): \textbf{Identity Leakage in Hairstyle}}, where reference identity features distort the source person.
    \textbf{(b): \textbf{Flaw Leakage in Bald}}, where flaws in the ``bald'' image pollute the hair representation and misleads the generative model to produce a failed hair contour.
    }
\label{fig:introduction_figure}
\vspace*{-3mm}
\end{figure}

\section{Introduction}
\label{sec:introduction}
In real-world scenarios such as hair salons and wig stores, customers often wish to preview how they would look with a new hairstyle by indicating a reference photo of a desired model. This motivates the task of hairstyle transfer, which aims to synthesize a realistic portrait by transplanting the hairstyle from a reference image onto the source person while preserving the source’s identity. However, constrained by the complexity of hair structure and texture---a scenario where the reference image is important for guiding the synthesis---achieving high-fidelity and controllable hairstyle transfer remains a significant challenge. While recent large-scale foundation models (e.g., Nano Banana Pro, Flux.2 Pro) have revolutionized image generation~\cite{li2026iv}, they fundamentally struggle with the zero-shot disentanglement required for precise local editing. When prompted to transfer a hairstyle, these models inherently entangle the reference hair with its original facial identity, pose, or lighting, failing to provide explicit control. To address this, the dominant paradigm for this task~\cite{zhu2022hairnet,Wu_2022_CVPR,wei2023hairclipv2,zhang2025stable,sun2025stable} follows a two-stream design to force spatial and semantic disentanglement: a geometric stream derived from the source (typically a ``bald'' image) and a hairstyle stream extracted from the reference. These two streams are fused by a synthesis backbone, often implemented as a diffusion model, to generate the final portrait. While this framework is intuitive and effective, this paradigm still suffers from artifacts such as identity inconsistency and mismatched hair geometry.

We demonstrate that these artifacts stem from a fundamental leakage problem, which is twofold: The first, termed \textit{Identity Leakage in Hairstyle} (Figure~\ref{fig:introduction_figure}(a)), arises when hairstyle features remain entangled with the reference’s identity and pose, leading to identity inconsistency in the synthesized face. The second, a more subtle phenomenon that we identify as \textit{Flaw Leakage in Bald} (Figure~\ref{fig:introduction_figure}(b)), occurs when generating the bald image from the source, the hair is typically not fully removed and a subtle flaw seen as shadow is left in the bald image. In this case, the flaw pollutes the hair representation and misleads the generative model to produce a failed hair contour. These two aspects represent complementary failure modes that collectively degrade the quality of the final synthesis. Specifically, the top row vividly demonstrates \textit{Identity Leakage in Hairstyle} where the synthesized face incorrectly inherits facial features from the reference; conversely, the bottom row showcases \textit{Flaw Leakage in Bald} where the model exploits subtle artifacts in the ``bald'' input, leading to a spurious correlation that erroneously replicates the source's original hairstyle. This limitation underscores the need for a framework that explicitly and jointly purifies both streams.

To this end, we propose the \textbf{Dual-Purification Framework (DPF)}. It introduces two specialized strategies to realize this goal: \textbf{Adversarial Hairstyle Purification (AHP)} explicitly suppresses identity information from hairstyle features by adversarially minimizing hairstyle–bald mutual information, where the discriminator serves as an implicit mutual information estimator that guides the generator to suppress identity-related cues between hairstyle and bald representations, while \textbf{Contrastive Geometric Purification (CGP)} suppresses \textit{Flaw Leakage} in the geometric stream by introducing a contrastive objective that penalizes the model's reliance on geometric flaws. By explicitly purifying both the hairstyle and geometric information streams, DPF achieves state-of-the-art performance in high-fidelity, identity-preserving hairstyle transfer.  Our contributions are summarized as follows:

\begin{itemize}
\item We systematically identify and formalize a fundamental leakage problem in hairstyle transfer. We further define its two aspects: \textit{Identity Leakage in Hairstyle}, which arises from feature entanglement, and \textit{Flaw Leakage in Bald}, which stems from spurious correlation learning.

\item We introduce the Dual Purification Framework (DPF), a novel architecture based on an explicit purification mechanism. DPF integrates two targeted modules (AHP and CGP) to explicitly and jointly purify the hairstyle and geometric information streams, thereby suppressing both Identity Leakage and Flaw Leakage.

\item Extensive experiments demonstrate that our framework sets a new state-of-the-art in high-fidelity and identity-preserving hairstyle transfer, outperforming prior methods both qualitatively and quantitatively.
\end{itemize}

\section{Related Work}

\noindent\textbf{Hairstyle Transfer.}
The evolution of hairstyle transfer has been largely driven by generative models. A significant line of work operates in the latent space of pretrained StyleGANs~\cite{Karras2018ASG,Karras2019AnalyzingAI,Karras2021Advances}, facing a trade-off between slow, high-fidelity optimization-based approaches such as Barbershop~\cite{zhu2021barbershop} and fast but less accurate encoder-based methods like HairFastGAN~\cite{nikolaev2024hairfastgan} and HairCLIP~\cite{wei2022hairclip}. However, these methods remain significantly constrained by the inherent entanglement in GAN feature spaces and often fail on out-of-distribution poses~\cite{zhu2022hairnet}. To mitigate geometric inconsistencies, recent works have explored 3D-aware representations using NeRFs~\cite{mildenhall2021nerf} or meshes, as in HairNeRF~\cite{chang2023hairnerf} and DELTA~\cite{feng2023learning}, though at the cost of increased complexity and data demands~\cite{kim20233d,feng2023learning}. Recently, large foundation models have demonstrated unprecedented capabilities in end-to-end image editing. However, applying them directly to personalized hairstyle transfer often fails due to severe feature entanglement—altering the hair typically compromises the source's facial identity or pose in a zero-shot setting. To enforce strict spatial and semantic disentanglement, diffusion-based approaches~\cite{rombach2022high} adopt a ``programmatic disentanglement'' strategy that synthesizes hair onto hair-agnostic representations~\cite{Wu_2022_CVPR,chung2025preserve,li2025haireditor}, exemplified by HairFusion~\cite{chung2025preserve} and Stable-Hair~\cite{zhang2025stable,sun2025stable}. Yet, our systematic diagnosis reveals a fundamental leakage problem in this paradigm, where the multi-stage condition fusion process propagates both identity features and geometric flaws into the final synthesis. We address this problem with a dual-purification framework that jointly purifies the hairstyle representation and the geometric pathway.

\noindent\textbf{Disentangled Representation Learning.}
Disentangled representation learning (DRL) aims to separate independent factors of variation in learned features~\cite{bengio2013representation,Higgins2016betaVAELB,wang2024disentangled,lee2018diverse}. Typical methods train encoders from scratch to be invariant to nuisance factors via adversarial or contrastive objectives~\cite{ganin2015unsupervised,kim2018disentangling}, and have been widely applied in facial attribute editing~\cite{liu2018unified,deng2020disentangled,liu2021afd,huang2023adaptive,ren2023reinforced,dalva2023face,pang2025disHPL}. However, such models learn disentanglement solely from raw data. In contrast, our work leverages general-purpose foundation models such as DINOv2~\cite{oquab2023dinov2} and CLIP~\cite{radford2021learning,dai2023instructblip} as rich feature sources. Despite their robustness, these representations remain highly entangled for fine-grained tasks~\cite{zhou2022retrieval,zhou2025perception,li2024uncertaintyrag}. Instead of using them for external supervision~\cite{liang2025differ,sun2023eva}, we propose a \textit{post-hoc} framework that directly purifies these features—disentangling identity information while mitigating the model's reliance on irrelevant data artifacts to achieve effective disentanglement without costly retraining.

\begin{figure*}[t]
	\centering
	\includegraphics[width=\textwidth]{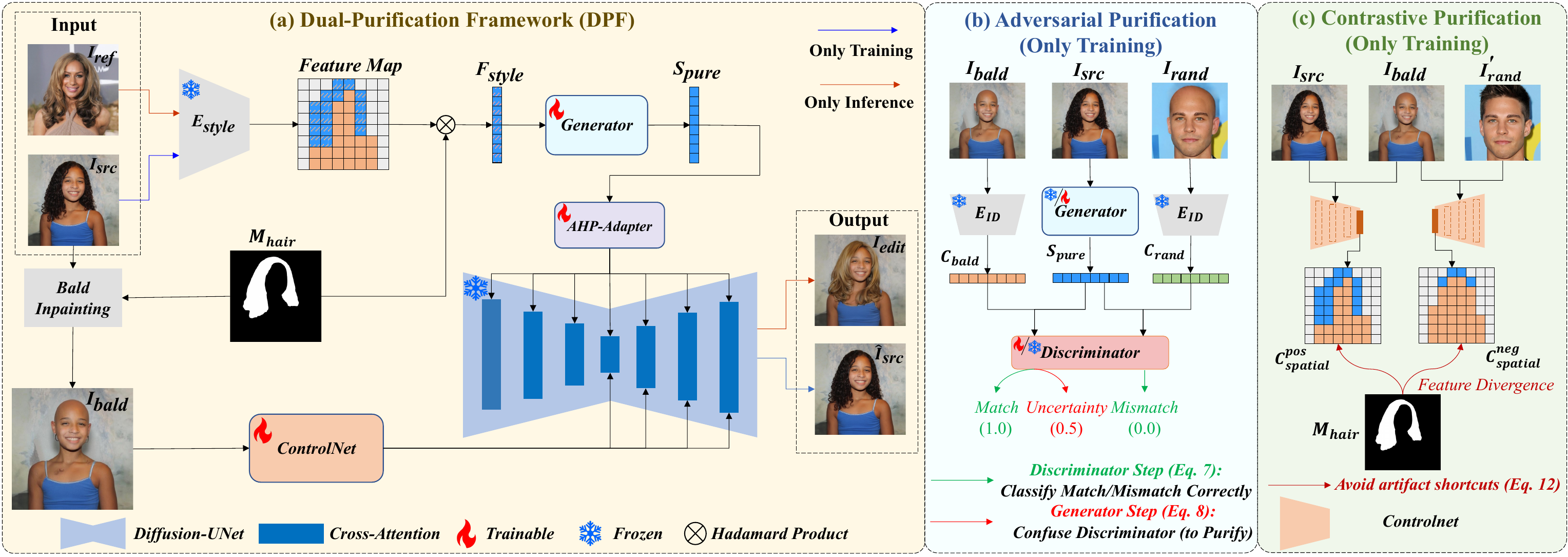}
	\vspace*{-5mm}
	\caption{
		\textbf{The Dual-Purification Framework (DPF).}
		\textbf{(a) Simplified Pipeline Overview.} This diagram illustrates the core data flow, which is shared by both the inference process and the training's self-reconstruction pass. As shown in the top-right corner, the data flow involves a sequence of modules for feature extraction and synthesis. 
		\textbf{(b) Adversarial Hairstyle Purification (AHP):} To resolve \textit{Identity Leakage in Hairstyle}, the Generator is trained to fool a Discriminator which learns to detect identity information within the style representation \(S_{pure}\).
		\textbf{(c) Contrastive Geometric Purification (CGP):} To resolve \textit{Flaw Leakage in Bald}, a contrastive loss forces the ControlNet to produce divergent features for different styles, suppressing the reliance on artifacts from \(I_{bald}\) by penalizing their use.
		\textbf{Note that modules (b) and (c) act as explicit regularizers utilized exclusively during the training phase, adding no computational overhead during inference.}}
	\label{fig:framework}
\end{figure*}  
\section{Methodology}
\subsection{Preliminaries}
\label{sec:preliminaries}
Our work builds on the Latent Diffusion Model (LDM)~\cite{rombach2022high} framework. We integrate two key conditioning mechanisms: \textit{ControlNet}~\cite{zhang2023adding} for fine-grained spatial control, and an IP-Adapter-inspired \cite{ye2023ip} module \textit{AHP-Adapter} for style injection. While the \textit{AHP-Adapter} provides a powerful mechanism for style conditioning, its direct use of raw image features leads to the \textit{Identity Leakage in Hairstyle} problem, as the features remain highly entangled. Concurrently, \textit{ControlNet}, which relies on pre-processed inputs like the $I_{bald}$ condition, introduces a complementary challenge: these inputs can contain subtle artifacts from their generation (e.g., inpainting), which the model can exploit as a ``shortcut'', leading to the \textit{Flaw Leakage in Bald} failure mode we identified in the introduction.

Our work directly addresses these shortcomings by proposing a \textbf{Dual-Purification Framework (DPF)} to purify both the style features (via \textbf{AHP}) and the geometric condition (via \textbf{CGP}). For a detailed background on these foundational models, including their original formulations, please refer to the Appendix.

\subsection{Overall Architecture}
\label{sec:overall_architecture}
Our DPF, illustrated in Figure~\ref{fig:framework}, is trained via a self-reconstruction task. Given a source image $I_{src}$ and its corresponding hair mask $M_{hair}$, we first generate a ``bald'' portrait $I_{bald}$ using a specialized diffusion-based inpainting model (detailed in the Appendix). The framework then synthesizes a reconstruction $\hat{I}_{src}$ through two parallel pathways.

\noindent\textbf{Style Pathway.} A frozen style encoder $E_{style}$ extracts patch features from the source image. These are masked to produce a variable-length sequence of entangled hair features $F_{style} \in \mathbb{R}^{N \times D_{in}}$.
\begin{equation}
F_{style} = E_{style}(I_{src}) \otimes M_{hair}
\end{equation}
Our trainable style generator $G$ then purifies these features into a fixed-length, disentangled style representation $S_{pure} \in \mathbb{R}^{k \times D_{out}}$:
\begin{equation}
S_{pure} = G(F_{style})
\label{eq:style_purification}
\end{equation}

\noindent\textbf{Geometric Pathway.} Concurrently, a trainable \textit{ControlNet} $CNet$ and a frozen identity encoder $E_{ID}$ extract spatial and identity conditions from the bald portrait:
\begin{equation}
    C_{spatial} = CNet(I_{bald}, S_{pure})
\label{eq:geom_conditions}
\end{equation}
where $C_{spatial}$ provides multi-level geometric guidance. In our implementation, $E_{ID}$ and $E_{style}$ share the same pre-trained encoder, where $E_{ID}$ extracts the global [CLS] token and $E_{style}$ extracts the patch-level features.

\noindent\textbf{Synthesis.} Finally, a pre-trained latent diffusion model (LDM) backbone, $\mathcal{G}_{LDM}$, generates the edited image conditioned on both pathways:
\begin{equation}
\label{eq:synthesis}
I_{edit} = \mathcal{G}_{LDM}(C_{spatial}, S_{pure})
\end{equation}

\noindent\textbf{Dual-target Self-reconstruction Training.}
Our DPF is trained via a \textit{dual-target self-reconstruction} task. While traditional methods rely on low-quality video or synthetic datasets, our framework maintains the advantage of single-image self-supervision, requiring only $I_{src}$ and its derived $I_{bald}$ for training. However, to explicitly establish the geometric anchor and prevent shortcut learning, we adaptively switch the training target based on the style condition. Specifically, in the conditional pass where $S_{pure}$ is active, the model minimizes the reconstruction loss ($\mathcal{L}_{rec}$) between $I_{src}$ and the synthesized image $\hat{I}_{src}$ to learn texture and fidelity (blue flow in Fig~\ref{fig:framework} (a)). Conversely, in the 10\% unconditional pass where $S_{pure}$ is dropped, the training target switches from $I_{src}$ to the bald portrait $I_{bald}$. This dual-target design forces the unconditional branch $\epsilon_{uncond}$ to function as a pure geometric baseline, preventing the model from ``leaking'' original hair features from $I_{bald}$'s artifacts. Despite this target switching, the training remains strictly within the self-reconstruction paradigm as all supervision signals are derived from the same source subject, naturally conferring robust cross-sample transfer capability during inference.

\subsection{Adversarial Purification of Hairstyle Features}
\label{sec:ahp}

The cornerstone of our framework is the \textbf{Adversarial Hairstyle Purification (AHP)} module, illustrated in Figure~\ref{fig:framework}(b). Its objective is to transform the entangled hairstyle features $F_{style}$ into a purified representation $S_{pure}$ that is disentangled from identity and pose. To achieve this, we formulate an adversarial game between a lightweight style generator $G$ and a content-aware discriminator $D$.

The generator maps the variable-length hairstyle features $F_{style}\in\mathbb{R}^{N\times D_{in}}$ to a fixed-length purified representation
\begin{equation}
	S_{pure}=G(F_{style}), \qquad
	S_{pure}\in\mathbb{R}^{k\times D_{out}},
\end{equation}
while the discriminator evaluates the compatibility between $S_{pure}$ and an identity representation. Detailed architectures of $G$ and $D$ are provided in the Appendix. During training, the discriminator receives positive pairs$(S_{pure},C_{bald})$, where both features originate from the same identity, and negative pairs $(S_{pure},C_{rand})$, where the identity embedding is randomly sampled from another image. Denoting these two distributions by
$\mathbf z\sim\mathcal P_{pos}$ and
$\mathbf z'\sim\mathcal P_{neg}$,
inspired by mutual information (MI) minimization~\cite{Belghazi2018MutualIN}, the adversarial objective encourages the generator to reduce identity predictability from the purified hairstyle representation. Ideally, this corresponds to reducing the dependence between $S_{pure}$ and $C_{bald}$,
\begin{equation}
	\label{eq:mi_minimization}
	\min_G \operatorname{MI}(S_{pure};C_{bald}).
\end{equation}
Intuitively, if the discriminator can no longer infer whether a hairstyle feature matches a given identity, the hairstyle representation contains little identity information, resulting in a more disentangled representation. We therefore use the discriminator as an adversarial proxy for this MI-inspired purification objective. The discriminator is optimized using the standard binary cross-entropy objective,
\begin{equation}
	\label{eq:loss_d}
	\begin{split}
		\mathcal L_D
		=
		-\mathbb E_{\mathbf z\sim\mathcal P_{pos}}
		[\log D(\mathbf z)]
		-
		\mathbb E_{\mathbf z'\sim\mathcal P_{neg}}
		[\log(1-D(\mathbf z'))].
	\end{split}
\end{equation}
Unlike conventional adversarial learning, the generator does not attempt to flip the discriminator's prediction. Instead, its objective is to make positive pairs indistinguishable from negative ones, driving the discriminator toward maximal uncertainty. Therefore, the generator minimizes
\begin{equation}
	\label{eq:loss_g_adv}
	\mathcal L_{adv}
	=
	\mathbb E_{\mathbf z\sim\mathcal P_{pos}}
	\left[
	\left(D(\mathbf z)-0.5\right)^2
	\right].
\end{equation}
This objective fundamentally differs from the non-saturating GAN loss~\cite{goodfellow2014generative}. Rather than maximizing the discriminator's confidence to imitate a target distribution, our objective drives the discriminator toward maximal uncertainty, thereby removing identity-related information from the hairstyle representation instead of generating more realistic features.

\subsection{Contrastive Geometric Purification (CGP)}
\label{sec:cgp}

While AHP (Sec.~\ref{sec:ahp}) purifies the style representation, it does not address the second critical failure mode: \textit{Flaw Leakage in Bald}. Our analysis reveals that this leakage primarily originates within the conditional \textit{ControlNet}. Tasked with providing geometric guidance from the $I_{bald}$ input, the \textit{ControlNet} tends to overfit subtle algorithmic artifacts (e.g., remnants of the original hair boundary), leading to shortcut learning~\cite{geirhos2020shortcut,ilyas2019adversarial}. Consequently, the conditional features become dominated by these artifacts rather than being properly guided by the hairstyle representation $S_{pure}$.

\noindent\textbf{Style-Sensitive Contrastive Objective.}
To suppress this shortcut, we employ a style-sensitive contrastive learning scheme during training. For each sample, the same geometric condition $I_{bald}$ is paired with two contrasting hairstyle conditions: the target style representation $S_{pure}$ and a negative style representation $S_{neg}$ extracted from another image $I'_{rand}$. The corresponding multi-scale ControlNet features are computed as
\begin{equation}
	C^{pos}=CNet(I_{bald},\,S_{pure}),
	\label{eq:cnet_pos}
\end{equation}
\begin{equation}
	C^{neg}=CNet(I_{bald},\,S_{neg}),
	\label{eq:cnet_neg}
\end{equation}
where $S_{neg}$ is extracted using the same hairstyle encoder from a randomly sampled image $I'_{rand}$. The complete construction of the contrastive objective is illustrated in Figure~\ref{fig:framework}(c).

\begin{algorithm}[t]
\caption{Incremental CFG for DPF Inference}
\label{alg:incremental_cfg}
\textbf{Input:} Noisy latent $\mathbf{z}_t$, Timestep $t$, Geometric condition $C_{spatial}$, Purified style $S_{pure}$, Guidance scale $\omega$ \\
\textbf{Output:} Predicted noise $\hat{\boldsymbol{\epsilon}}$
\begin{algorithmic}[1]
\item[] \textit{\% Step 1: Unconditional pass (Bald geometry anchor)}
\STATE $\boldsymbol{\epsilon}_{uncond} \leftarrow \epsilon_\theta(\mathbf{z}_t, t, C_{spatial}, \emptyset)$
\item[] \textit{\% Step 2: Conditional pass (Target hairstyle injection)}
\STATE $\boldsymbol{\epsilon}_{cond} \leftarrow \epsilon_\theta(\mathbf{z}_t, t, C_{spatial}^{pos}, S_{pure})$
\item[] \textit{\% Step 3: Incremental extrapolation}
\STATE $\hat{\boldsymbol{\epsilon}} \leftarrow \boldsymbol{\epsilon}_{uncond} + \omega \cdot (\boldsymbol{\epsilon}_{cond} - \boldsymbol{\epsilon}_{uncond})$
\STATE \textbf{return} $\hat{\boldsymbol{\epsilon}}$
\end{algorithmic}
\end{algorithm}

To discourage this shortcut behavior, CGP enforces the two mid-block activations to be dissimilar within the hair region. We first compute the mean spatial cosine similarity $\text{Sim}_{hair}$ between $C^{pos}_{spatial}$ and $C^{neg}_{spatial}$, evaluated only inside the ground-truth hair mask $M_{hair}$:
\begin{equation}
\label{eq:sim_cgp}
\text{Sim}_{hair} = 
\frac{\sum_{i,j} 
\cos\!\left(\bar{C}^{neg,(i,j)}_{spatial},\, \bar{C}^{pos,(i,j)}_{spatial}\right)
\otimes M_{hair}^{i,j}}
{\sum_{i,j} M_{hair}^{i,j}},
\end{equation}
where $\bar{C}$ denotes feature maps that have been explicitly processed by Instance Normalization (IN) followed by L2-normalization, and $(i,j)$ indexes the spatial positions. The inclusion of Instance Normalization is a crucial design choice: it strips away global style information such as color and illumination distributions from the activations, strictly isolating the pure geometric and structural contours. 
Because the features now represent pure geometry, the spatial contours of two distinct hairstyles should be nearly orthogonal. To prevent over-penalization and stabilize training when the geometric features are already sufficiently divergent, we introduce a similarity margin $\tau$ (detailed implementation can be found in Appendix). The final CGP loss $\mathcal{L}_{CGP}$ is defined as:
\begin{equation}
\label{eq:loss_cgp}
\mathcal{L}_{CGP} = \max(0, \text{Sim}_{hair} - \tau).
\end{equation}
As illustrated in Figure~\ref{fig:framework}(c), to ensure that CGP does not interfere with the primary reconstruction pathway, gradients from $\text{Sim}_{hair}$ are propagated only through the negative branch ($C^{neg}_{spatial}$), while the positive branch ($C^{pos}_{spatial}$) is detached. Minimizing $\mathcal{L}_{CGP}$ directly penalizes the geometric similarity between the two activations when it exceeds the margin $\tau$. This explicitly compels the \textit{ControlNet} to inject accurate structural features from the target style rather than exploiting residual geometric artifacts in $I_{bald}$. CGP is applied exclusively during training.

\subsection{DPF Image Synthesis}
\label{sec:synthesis}

\noindent\textbf{Final Training Objective.}
The full DPF framework is trained by optimizing a joint objective. We reformulate the standard LDM reconstruction loss $\mathcal{L}_{rec}$ to align with our incremental training logic:
\begin{equation}
\label{eq:loss_rec_updated}
\mathcal{L}_{rec} = \mathbb{E}_{\mathbf{z}_0, \boldsymbol{\epsilon}, t} \left[ \left\| \boldsymbol{\epsilon} - \epsilon_\theta(\mathbf{z}_t, t, C_{spatial}, \hat{S}) \right\|^2_2 \right],
\end{equation}
where the training target $\mathbf{z}_0$ and style condition $\hat{S}$ are adaptively selected. With probability 0.9, $\hat{S} = S_{pure}$ and $\mathbf{z}_0 = E_{vae}(I_{src})$; with probability 0.1, $\hat{S} = \emptyset$ and $\mathbf{z}_0 = E_{vae}(I_{bald})$. This ensures that the unconditional pass $\epsilon_{uncond}$ acts as a pure geometric baseline. The final objective combined with AHP and CGP is:
\begin{equation}
\label{eq:loss_total}
\mathcal{L}_{total} = \mathcal{L}_{rec} + \lambda_{adv} \mathcal{L}_{adv} + \lambda_{CGP} \mathcal{L}_{CGP}.
\end{equation}

\noindent\textbf{Inference via Incremental CFG.}
During inference, we introduce an \textit{incremental} Classifier-Free Guidance (CFG) strategy, summarized in Algorithm~\ref{alg:incremental_cfg}. The core of this strategy lies in its synergy with our dual-target training: by training the unconditional branch $\epsilon_{uncond}$ to specifically reconstruct the bald portrait $I_{bald}$ (Sec.~\ref{sec:overall_architecture}), we establish it as a stable \textit{geometric baseline anchor}. Like recent image-to-image translation frameworks~\cite{brooks2023instructpix2pix}, our unconditional pass explicitly retains the spatial condition $C_{spatial}$ while dropping only the style ($\hat{S}=\emptyset$). Consequently, the predicted increment $\Delta\epsilon = (\epsilon_{cond} - \epsilon_{uncond})$ represents a \textit{purified hairstyle increment}. Since both branches share the same geometric anchor and $\epsilon_{uncond}$ is a faithful representation of the bald identity, the delta $\Delta\epsilon$ ideally remains zero in non-hair regions (e.g., face and skin). This property allows the final synthesis, extrapolated via the guidance scale $\omega$, to inject rich hairstyle textures while ensuring precise structural alignment and zero skin-tone drift.

\begin{table}[t]
	\centering
	\caption{
		\textbf{Encoder Alignment with Perceptual Dimensions.}
		Spearman's correlation (\(\rho\)) measures the alignment of encoder feature distances with perceptual \textbf{Texture} and \textbf{Structure}. Higher \(\rho\) is better. The results reveal a clear trade-off: no single encoder excels at both, motivating our hybrid design.
	}
	\label{tab:decomposed_analysis}
	\resizebox{0.95\linewidth}{!}{%
		\begin{tabular}{llcc}
			\toprule
			\textbf{Encoder Model} & \textbf{Paradigm} & \textbf{Texture} ($\rho \uparrow$) & \textbf{Structure} ($\rho \uparrow$) \\
			\midrule
			BiSeNet~\cite{yu2018bisenet} & Task-Specific  & -0.265 & 0.563 \\
			Qwen2.5VL-ViT~\cite{bai2025qwen2} & Large VLM & -0.277 & 0.459 \\
			\midrule
			DINOv2~\cite{oquab2023dinov2} & Intra-modal & -0.370 & \textbf{0.632} \\
			ResNet-50~\cite{he2016deep} & Supervised & -0.364 & 0.623 \\
			\midrule
			CLIP-L/14~\cite{radford2021learning} & \multirow{2}{*}{Cross-modal} & 0.109 & 0.290 \\
			CLIP-B/16~\cite{radford2021learning} & & \textbf{0.220} & 0.150 \\
			\bottomrule
		\end{tabular}%
	}
\end{table}

\begin{figure*}[t] 
    \centering 
    \begin{subfigure}{\linewidth}
        \centering
        \includegraphics[width=\linewidth]{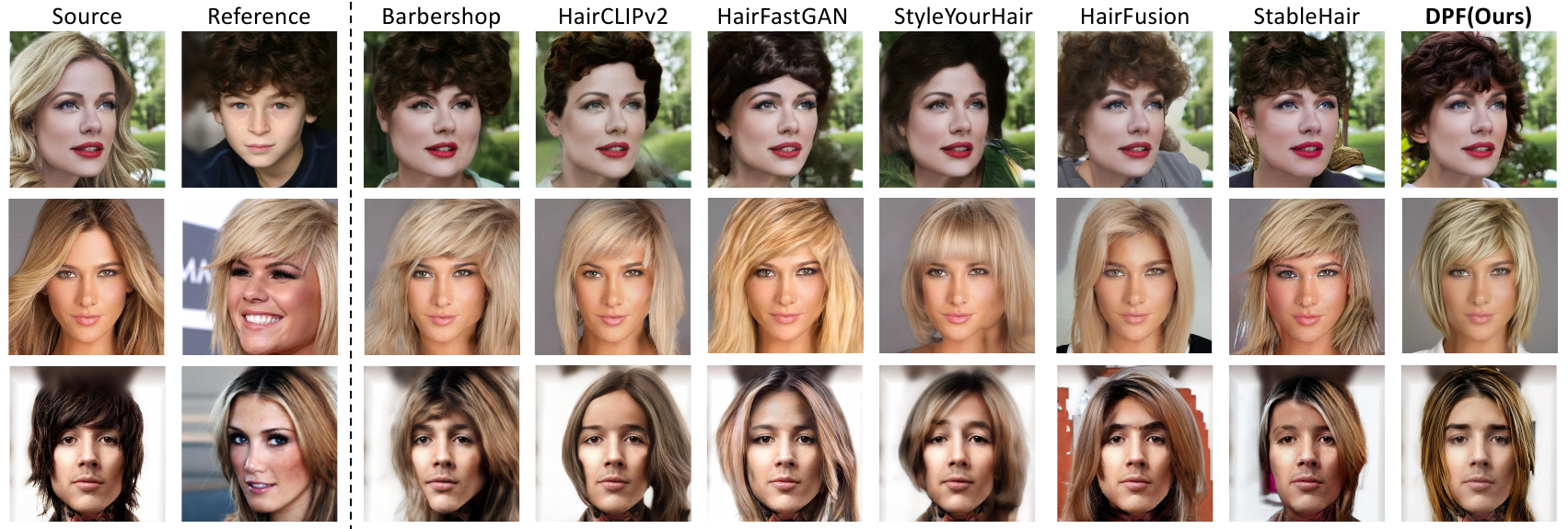}
        \vspace*{-5mm}
        \caption{Results on CelebA-HQ benchmark~\cite{karras2017progressive}}
        \vspace*{2mm}
        \label{fig:qual_celebahq_sub}
    \end{subfigure}
    \begin{subfigure}{\linewidth}
        \centering
        \includegraphics[width=0.85\linewidth]{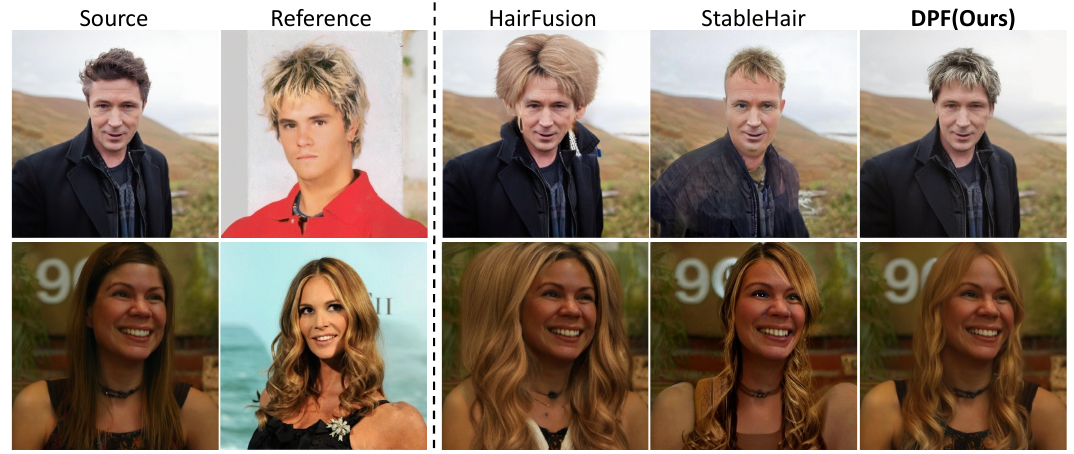}
        \vspace*{-2mm}
        \caption{Results on Human-Matting benchmark~\cite{chen2018semantic}}
        \label{fig:qual_humanmatting_sub}
		\vspace*{-2mm}
    \end{subfigure}
    \caption{
        \textbf{Qualitative results of DPF on diverse benchmarks.}
        \textbf{(a) CelebA-HQ~\cite{karras2017progressive}:} On standard portraits, DPF generates photorealistic results, mitigating both \textit{Identity Leakage in Hairstyle} and \textit{Flaw Leakage in Bald}.
        \textbf{(b) Human-Matting~\cite{chen2018semantic}:} On these challenging in-the-wild images, DPF demonstrates superior robustness and generation quality compared to prior work.
    }
    \label{fig:qualitative_results_combined}

\end{figure*}
  
\section{Experiments}
\label{sec:experiments}
\begin{table*}[t]
\centering 

\caption{
    \textbf{Quantitative comparison with state-of-the-art methods on diverse benchmarks.}
    Our DPF demonstrates superior or highly competitive performance across most standard metrics on both high-quality portraits and challenging in-the-wild datasets. Best results are in \textbf{bold}, second best are \underline{underlined}.
}
\vspace*{-3mm}
\label{tab:sota_combined}

\def\subtablewidth{0.85\textwidth}

\subcaption{Results on the CelebA-HQ benchmark~\cite{karras2017progressive}}
\vspace*{-2mm}
\label{tab:sota_celebahq_sub}
\begin{tabularx}{\subtablewidth}{l c >{\centering\arraybackslash}X >{\centering\arraybackslash}X >{\centering\arraybackslash}X >{\centering\arraybackslash}X >{\centering\arraybackslash}X}
    \toprule
    \multirow{2}{*}{\textbf{Method}} & \multirow{2}{*}{\textbf{Architecture}} & 
    \textbf{Identity} & \multicolumn{2}{c}{\textbf{Reconstruction}} & \multicolumn{2}{c}{\textbf{Fidelity}} \\
    \cmidrule(lr){3-3} \cmidrule(lr){4-5} \cmidrule(lr){6-7} 
    & & \textbf{ID-Sim} $\uparrow$ & \textbf{SSIM\textsubscript{nh}} $\uparrow$ & \textbf{PSNR\textsubscript{nh}} $\uparrow$ & \textbf{FID} $\downarrow$ & \textbf{H-LPIPS} $\downarrow$ \\
    \midrule
    Barbershop~\cite{zhu2021barbershop} & GAN & 0.73 & \textbf{0.87} & 14.10 & 26.69 & \textbf{0.306} \\
    StyleYourHair~\cite{kim2022style} & GAN & 0.75 & \underline{0.86} & 15.27 & 23.02 & 0.367 \\
    HairCLIPv2~\cite{wei2023hairclipv2} & GAN & \underline{0.79} & 0.85 & \textbf{17.06} & 17.59 & 0.394 \\
    HairFastGAN~\cite{nikolaev2024hairfastgan} & GAN & 0.78 & 0.82 & 15.73 & \underline{17.27} & 0.381 \\
    StableHair~\cite{zhang2025stable} & Diffusion & 0.78 & 0.84 & 16.78 & 29.81 & 0.394 \\
    HairFusion~\cite{chung2025preserve} & Diffusion & 0.75 & \underline{0.86} & 15.79 & 40.24 & 0.410 \\
    \midrule
    \textbf{DPF (Ours)} & Diffusion & \textbf{0.80} & \textbf{0.87} & \underline{16.88} & \textbf{16.53} & \underline{0.356} \\
    \bottomrule
\end{tabularx}

\vspace{2mm} 

\subcaption{Results on the Human-Matting benchmark~\cite{chen2018semantic}}
\vspace*{-2mm}
\label{tab:sota_humanmatting_sub}
\begin{tabularx}{\subtablewidth}{l c >{\centering\arraybackslash}X >{\centering\arraybackslash}X >{\centering\arraybackslash}X >{\centering\arraybackslash}X >{\centering\arraybackslash}X}
    \toprule
    \multirow{2}{*}{\textbf{Method}} & \multirow{2}{*}{\textbf{Architecture}} & 
    \textbf{Identity} & \multicolumn{2}{c}{\textbf{Reconstruction}} & \multicolumn{2}{c}{\textbf{Fidelity}} \\
    \cmidrule(lr){3-3} \cmidrule(lr){4-5} \cmidrule(lr){6-7}
    & & \textbf{ID-Sim} $\uparrow$ & \textbf{SSIM\textsubscript{nh}} $\uparrow$ & \textbf{PSNR\textsubscript{nh}} $\uparrow$ & \textbf{FID} $\downarrow$ & \textbf{H-LPIPS} $\downarrow$ \\
    \midrule
    StableHair~\cite{zhang2025stable} & Diffusion & 0.67 & \underline{0.84} & \underline{16.14} & 16.45 & \underline{0.468} \\
    HairFusion~\cite{chung2025preserve} & Diffusion & \underline{0.70} & 0.83 & 15.65 & \underline{15.81} & 0.504 \\
    \midrule
    \textbf{DPF (Ours)} & Diffusion & \textbf{0.75} & \textbf{0.87} & \textbf{16.98} & \textbf{7.75} & \textbf{0.417} \\
    \bottomrule
\end{tabularx}

\end{table*}

\subsection{Encoder Analysis}
\label{sec:encoder_analysis}

A powerful hairstyle representation must capture both perceptual texture (e.g., color, gloss) and geometric structure (e.g., shape, volume). To identify the best pre-trained encoders for these distinct roles, we conduct a decomposed analysis. We measure the alignment between an encoder's feature space and these two perceptual dimensions by computing two separate Spearman's rank correlations ($\rho$). We correlate the feature distances against a perceptual texture metric (e.g., Masked LPIPS) and a geometric structure metric (e.g., Mask IoU), with full methodological details in the Appendix.

Our results, summarized in Table~\ref{tab:decomposed_analysis}, reveal a critical trade-off inherent in modern pre-training paradigms. We find that cross-modal models like \textit{CLIP}~\cite{radford2021learning} excel at capturing texture ($\rho=0.220$) but are poor at discerning structure. Conversely, self-supervised vision models like \textit{DINOv2}~\cite{oquab2023dinov2} are superior for geometric alignment ($\rho=0.632$) but exhibit a negative correlation with texture. 

We observe that none of the tested encoders excels at both, which directly motivates our hybrid approach. Instead of relying on a single, compromised encoder, our framework is designed to harness the complementary strengths of a \textit{``texture champion''} (CLIP) and a \textit{``structure champion''} (DINOv2). This analysis provides strong empirical support for this design choice in DPF. The following sections will now present a comprehensive evaluation, demonstrating how this principled approach enables state-of-the-art performance in hairstyle transfer.

\begin{table*}[!t]
\centering
\def\ablationwidth{0.85\textwidth}

\caption{
    \textbf{Ablation Study of DPF Components.} 
    We progressively build our full DPF model to validate the contribution of each key design choice. 
    \textbf{ID-Sim} measures high-level identity preservation. 
    \textbf{SSIM\textsubscript{nh}} and \textbf{PSNR\textsubscript{nh}} measure the pixel-level fidelity of the \textit{non-hair regions} (face/background) against the source.
    \textbf{FID} measures realism, and \textbf{LFR} (Leakage Failure Rate) measures the \textit{Flaw Leakage in Bald} failure.
}
\vspace*{-2mm}
\label{tab:ablation}

\begin{tabularx}{\ablationwidth}{l l >{\centering\arraybackslash}X >{\centering\arraybackslash}X >{\centering\arraybackslash}X >{\centering\arraybackslash}X >{\centering\arraybackslash}X}
\toprule
& \textbf{Method Configuration} & \textbf{ID-Sim} $\uparrow$ & \textbf{SSIM\textsubscript{nh}} $\uparrow$ & \textbf{PSNR\textsubscript{nh}} $\uparrow$ & \textbf{FID} $\downarrow$ & \textbf{LFR} $\downarrow$ \\
\midrule
(1) & Baseline (IP-Adapter + CN) & 0.76 & \textbf{0.87} & 15.62 & 28.85 & 37.86\% \\
(2) & (1) + AHP & \textbf{0.80} & \underline{0.86} & \underline{16.55} & \underline{17.25}  & 35.19\% \\
(3) & (2) + CGP & \underline{0.79} & \underline{0.86} & 16.12 & 18.53 & \underline{10.33\%} \\
(4) & \textbf{Ours} ((3) + Hybrid Feat.) & \textbf{0.80} & \textbf{0.87} & \textbf{16.88} & \textbf{16.53} & \textbf{3.17\%} \\
\bottomrule
\end{tabularx}

\end{table*}

\subsection{Experimental Setup}
\label{sec:experimental_setup}
\textbf{Dataset and Benchmarks.}
To train a model that is both high-fidelity and robust, we curate a novel training dataset, named \textbf{WildBald-HQ} (60k images). It is constructed by balancing 30,000 high-quality, aligned portraits from FFHQ~\cite{Karras2019AnalyzingAI} with 30,000 diverse, in-the-wild half-body portraits from the Human-Matting dataset~\cite{chen2018semantic}. This balanced composition is specifically designed to endow our model with the ability to handle both ideal and challenging real-world conditions. For evaluation, we conduct a comprehensive analysis on two distinct benchmarks: the CelebA-HQ dataset~\cite{karras2017progressive} for the standard headshot task, and the validation set of HumanMatting for the more challenging in-the-wild setting. The detailed dataset sampling strategy is provided in the Appendix.

\noindent\textbf{Implementation.}
Our framework is built upon Stable Diffusion v1.5 and a \textit{ControlNet} that is initialized from the official inpainting model weights~\cite{zhang2023adding} and subsequently fine-tuned as part of our framework. Guided by our analysis in Sec.~\ref{sec:encoder_analysis}, we adopt a \textit{hybrid feature strategy} to construct our initial style representation. We leverage the complementary strengths of both the ``texture champion'', CLIP-B/16~\cite{radford2021learning}, and the ``structure champion'', DINOv2~\cite{oquab2023dinov2}. Both sets of features are processed in parallel by our Generator, and their output tokens are combined to form the final purified style representation. This approach provides our AHP module with a rich but deeply entangled set of features, creating an ideal and challenging testbed to validate our purification paradigm. The specific fusion method for these features is detailed in the Appendix. We compare our method against state-of-the-art baselines including Barbershop~\cite{zhu2021barbershop}, HairCLIPv2~\cite{wei2023hairclipv2}, HairFastGAN~\cite{nikolaev2024hairfastgan}, StyleYourHair~\cite{kim2022style}, HairFusion~\cite{chung2025preserve}, and StableHair~\cite{zhang2025stable}. We evaluate the generated results using standard metrics from four complementary aspects: ID-Similarity~\cite{deng2019arcface} evaluates identity preservation; SSIM\textsubscript{nh}~\cite{zhou2004image} and PSNR\textsubscript{nh} measure reconstruction quality in the non-hair regions; FID~\cite{heusel2017gans} evaluates overall image realism; and H-LPIPS~\cite{zhang2018unreasonable} measures hairstyle fidelity with respect to the reference hairstyle. Complete training details are provided in the Appendix.

\subsection{Comparison with State-of-the-Art Methods}

We conduct a comprehensive evaluation of our \textbf{Dual-Purification Framework (DPF)} against a wide range of \textit{state-of-the-art} hairstyle transfer methods on two challenging benchmarks: the high-quality headshot of CelebA-HQ~\cite{karras2017progressive} and the more diverse, unaligned half-body portraits from the Human-Matting dataset~\cite{chen2018semantic}.

\noindent\textbf{Qualitative Comparison.}
As shown in Figure~\ref{fig:qualitative_results_combined}, our qualitative results illustrate DPF's strong performance. Across both CelebA-HQ (Fig.~\ref{fig:qual_celebahq_sub}) and the more challenging Human-Matting benchmark (Fig.~\ref{fig:qual_humanmatting_sub}), prior methods frequently suffer from the two core failure modes we identified: \textit{Identity Leakage in Hairstyle}, which causes identity inconsistency, and \textit{Flaw Leakage in Bald}, which wrongly reconstructs the source hairstyle. In stark contrast, our DPF robustly generates photorealistic and geometrically precise results that faithfully preserve the source identity, even for complex hairstyles on subjects with varied poses and scales. This visually illustrates that by purifying both the style (via AHP) and geometric (via CGP) streams, DPF effectively mitigates the challenges of both \textit{Identity Leakage in Hairstyle} and \textit{Flaw Leakage in Bald}.

\noindent\textbf{Quantitative Comparison.} 
The quantitative results are summarized in Table~\ref{tab:sota_combined}. On the standard CelebA-HQ benchmark (Table~\ref{tab:sota_celebahq_sub}), DPF achieves state-of-the-art results in identity preservation (ID-Sim) and fidelity (FID), while maintaining highly competitive performance in reconstruction metrics. It is worth noting that since CelebA-HQ consists of perfectly aligned portraits, the performance of most baselines—particularly GAN-based models—has largely saturated in this domain. Even in this highly competitive scenario, DPF demonstrates superior identity preservation capability, outperforming representative GAN-based models such as HairCLIPv2 and the encoder-based HairFastGAN. The true strength of our dual-purification mechanism is more evident in the challenging Human-Matting dataset (Table~\ref{tab:sota_humanmatting_sub}). Unlike the constrained headshot setting, these in-the-wild images feature significant pose variations and complex backgrounds. In this regime, prior methods suffer from severe leakage problems (as analyzed in Table~\ref{tab:ablation}), leading to a significant drop across all metrics. In contrast, DPF maintains exceptional robustness, outperforming existing methods by a wide margin—for instance, achieving a more than 50\% reduction in FID from 15.81 to 7.75. This sharp contrast validates DPF's unique capability to handle complex leakage issues in real-world hairstyle transfer, successfully transforming a flawed geometric foundation into a reliable asset for synthesis.

\begin{table}[t]
  \centering
  \caption{
    \textbf{Comprehensive System and User Evaluation.}
    We evaluate system efficiency and user preference (1-5 scale).
  }
  \label{tab:user_study_efficiency}
  \vspace*{-2mm}
  \resizebox{0.95\columnwidth}{!}{%
  \begin{tabular}{l c c c c c} 
    \toprule
    \multirow{2}{*}{\textbf{Method}} & \multicolumn{3}{c}{\textbf{User Study Score} $\uparrow$} & \multirow{2}{*}{\textbf{Mem (GB)} $\downarrow$} & \multirow{2}{*}{\textbf{Time (s)} $\downarrow$} \\
    \cmidrule(lr){2-4}
    & \textbf{Style} & \textbf{ID} & \textbf{Real} & & \\
    \midrule
    HairFusion & 3.22 & 3.14 & 3.08 & $\sim$10.9 & $\sim$14 \\
    StableHair & 3.05 & 3.46 & 3.02 & $\sim$16.0 & \textbf{$\sim$6} \\
    \midrule
    \textbf{DPF (Ours)} & \textbf{3.75} & \textbf{3.92} & \textbf{3.82} & \textbf{$\sim$5.4} & \textbf{$\sim$6} \\
    \bottomrule
  \end{tabular}%
  }
\vspace*{-3mm}
\end{table}

\subsection{User Study and System Efficiency}
\label{sec:user_study}

To further validate DPF's applicability as an interactive multimedia system, we evaluate its human-centric performance and computational footprint. We conducted a blind user study ($N=22$) where participants rated the generated results on a scale of 1 to 5 based on three criteria: Style transfer accuracy, Identity (ID) preservation, and Realism. As shown in Table~\ref{tab:user_study_efficiency}, DPF achieves the highest user preference scores across all dimensions. Furthermore, DPF operates with a peak memory footprint of merely $\sim$5.4GB, making it highly accessible for consumer-grade GPUs and edge deployment, compared to the heavy resource demands of StableHair ($\sim$16.0GB) and HairFusion ($\sim$10.9GB). Crucially, DPF maintains a highly competitive synthesis time of approximately 6 seconds per image (using 30-step DPMSolver++), matching the speed of the fastest diffusion baseline while requiring significantly less VRAM. These results confirm that DPF provides a superior, resource-friendly, and efficient solution for real-world multimedia applications.

\subsection{Ablation Studies}
\label{sec:ablation}

\begin{figure}[t]
  \centering
  \includegraphics[width=0.8\columnwidth]{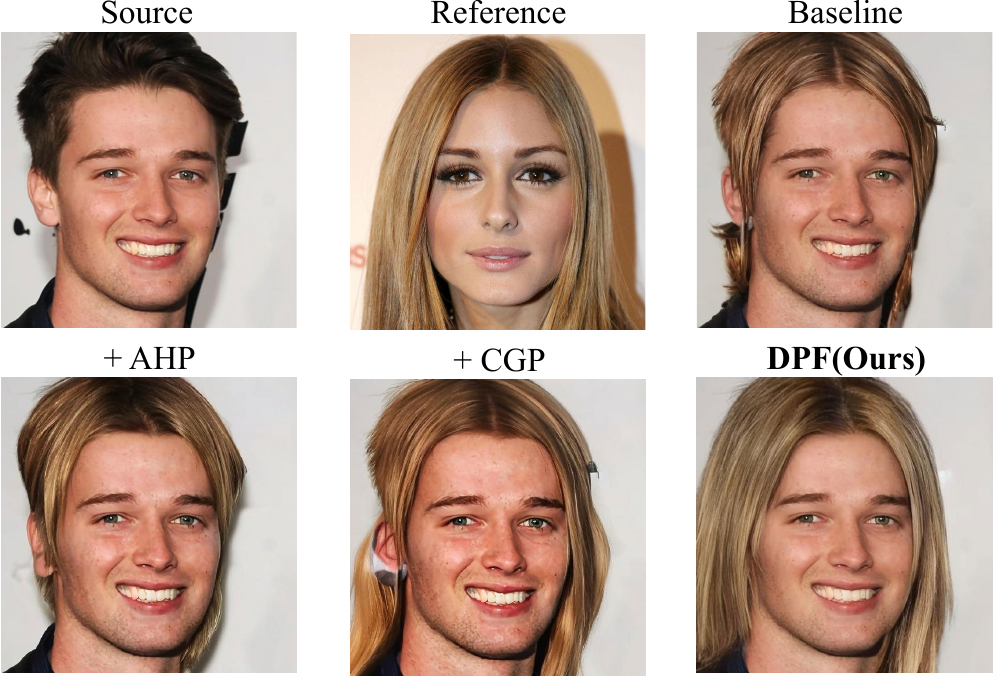}
  \vspace*{-4mm}
     \caption{
    \textbf{Qualitative Ablation of DPF Components.}
    Each step demonstrates the contribution of a key component.
}
\label{fig:visual_ablation}
\vspace*{-3mm}
\end{figure}

To validate the necessity and synergy of our proposed modules, we conduct a comprehensive ablation study, progressively building our full framework from a strong baseline. The quantitative results are summarized in Table~\ref{tab:ablation}, with corresponding qualitative results in Figure~\ref{fig:visual_ablation}.

Our analysis begins with the baseline (row 1), which struggles with identity preservation (low ID-Sim). Moreover, it suffers from frequent geometric failures. To quantify this, we define the \textit{Leakage Failure Rate (LFR)} as the percentage of cases where the Intersection over Union (IoU) between the source and generated hair masks exceeds 85\%. The baseline's high LFR indicates \textit{Flaw Leakage in Bald} is a significant problem. Integrating our AHP module (row 2) effectively addresses the identity problem, as shown by the sharp increase in ID-Sim score. However, AHP has a negligible effect on the leakage issue, with the LFR remaining high, demonstrating \textit{Flaw Leakage in Bald} is an orthogonal challenge. The subsequent step (row 3) introduces our CGP module. As validated by the results, CGP acts as an explicit regularizer to counter shortcut learning, drastically reducing the LFR from 35.19\% to 10.33\%. This indicates CGP is the key component for mitigating Flaw Leakage. However, as seen in Figure~\ref{fig:visual_ablation}, CGP exposes the poor structural quality of the baseline CLIP features (validated in Sec.~\ref{sec:encoder_analysis}). The final step (row 4) forms the full DPF (Ours) by introducing \textit{Hybrid Features}. This feature quality enhancement step provides the missing structural fidelity, mitigating remaining artifacts, achieving the best FID, and further suppressing the LFR to 3.17\%. This step-by-step analysis demonstrates the effectiveness of our DPF. It confirms that both purification modules are beneficial: AHP is crucial for mitigating \textit{Identity Leakage in Hairstyle}, while CGP is the key component for suppressing \textit{Flaw Leakage in Bald}. This synergy, combined with Hybrid Features for final structural fidelity, enables our high-quality results.

\section{Conclusion}
\label{sec:conclusion}

In this work, we identified a fundamental, twofold leakage problem in hairstyle transfer: \textit{Identity Leakage in Hairstyle} and \textit{Flaw Leakage in Bald}. To address this, we introduced the Dual-Purification Framework (DPF), an approach that integrates AHP to mitigate identity entanglement and CGP to suppress the critical shortcut learning causing flaw leakage. Our experiments demonstrate that this approach achieves state-of-the-art results, enabling robust synthesis despite the presence of geometric artifacts in the bald condition. The principles of DPF, such as \textit{explicit purification}, may be generalizable and offer a compelling direction for future research in other conditional generation domains.  
\begin{acks}
This work was supported in part by Beijing Natural Science Foundation L242092, Chinese National Natural Science Foundation Projects U23B2054, 62276254, 62206280, 62376265, the Beijing Science and Technology Plan Project Z231100005923033, the Science and Technology Development Fund of Macau Project 0140/2024/AGJ, and InnoHK program.
\end{acks}
\clearpage
\bibliographystyle{ACM-Reference-Format}
\bibliography{sample-base} 
\end{document}